\documentclass{article}

\usepackage[preprint]{neurips_2025}

\usepackage[utf8]{inputenc} % allow utf-8 input
\usepackage[T1]{fontenc}    % use 8-bit T1 fonts
\usepackage{hyperref}       % hyperlinks
\usepackage{url}            % simple URL typesetting
\usepackage{booktabs}       % professional-quality tables
\usepackage{amsfonts}       % blackboard math symbols
\usepackage{nicefrac}       % compact symbols for 1/2, etc.
\usepackage{microtype}      % microtypography
\usepackage{xcolor}         % colors
\usepackage{multirow}
\usepackage{booktabs}
\usepackage{multirow}
\usepackage{makecell}
\usepackage{tabularx}
\usepackage{soul}
\usepackage{amsmath}
\usepackage{siunitx}
\usepackage{graphicx} 
\usepackage{wrapfig}
\usepackage{tablefootnote}
\usepackage{subfigure}

\definecolor{mydarkblue}{rgb}{0,0.08,0.45}
\hypersetup{ %
    pdftitle={},
    pdfauthor={},
    pdfsubject={},
    pdfkeywords={},
    pdfborder=0 0 0,
    pdfpagemode=UseNone,
    colorlinks=true,
    linkcolor=mydarkblue,
    citecolor=mydarkblue,
    filecolor=mydarkblue,
    urlcolor=mydarkblue,
}

\newcommand\model{\text{XxaCT-NN }}
\newcommand{\highlight}[1]{\colorbox{blue!10}{#1}}

\title{XxaCT-NN: Structure Agnostic Multimodal Learning for Materials Science}

\author{%
  Jithendaraa Subramanian\\
  Toyota Research Institute\\
  Los Altos, CA 94022 \\
  \texttt{jith.subramanian@tri.global}\\
  \And
  Linda Hung\\
  Toyota Research Institute\\
  Los Altos, CA 94022 \\
  \texttt{linda.hung@tri.global}\\
  \And
  Daniel Schweigert\\
  Toyota Research Institute\\
  Los Altos, CA 94022 \\
  \texttt{daniel.schweigert@tri.global}\\
  \And
  Santosh Suram\\
  Toyota Research Institute\\
  Los Altos, CA 94022 \\
  \texttt{santosh.suram@tri.global}\\
  \And
  Weike Ye\\
  Toyota Research Institute\\
  Los Altos, CA 94022 \\
  \texttt{weike.ye@tri.global}\\
}

\begin{document}

\maketitle

\begin{abstract}

Recent advances in materials discovery have been driven by structure-based models, particularly those using crystal graphs. While effective for computational datasets, these models are impractical for real-world applications where atomic structures are often unknown or difficult to obtain. We propose a scalable multimodal framework that learns directly from elemental composition and X-ray diffraction (XRD)—two more available modalities in experimental workflows—without requiring crystal structure input. Our architecture integrates modality-specific encoders with a cross-attention fusion module and is trained on the 5-million-sample Alexandria dataset. We present masked XRD modeling (MXM), and apply MXM and contrastive alignment as self-supervised pretraining strategies. Pretraining yields faster convergence (up to 4.2× speedup) and improves both accuracy and representation quality. We further demonstrate that multimodal performance scales more favorably with dataset size than unimodal baselines, with gains compounding at larger data regimes. Our results establish a path toward structure-free, experimentally grounded foundation models for materials science.
\end{abstract}

\section{Introduction}
In large-scale materials models today, one of the most common representations of a material is through crystal graphs~\citep{xie2018crystal}. The crystal graph representation aligns with theoretical representations of materials at the atomic scale, and has enabled models to predict formation energy, stability, band gap, crystal structure, and more, in good agreement with computational materials datasets~\citep{reiser2022graph}. These models have also enabled the predicting of millions of new stable materials on computers~\citep{merchant2023scaling}.

However, when it comes to materials being made in a lab, details of the materials structure are often unknown or difficult to determine, which makes structural and crystal graph representations impractical for real materials discovery~\citep{montoya2024how}. For an experimentalist, information about material samples is instead made up of modalities like synthesis data -- the recipe used to synthesize the material -- and characterization (i.e. measurement) data. The majority of experimental data comes from characterizations, which includes techniques like X-ray diffraction (XRD), X-ray photoelectron spectroscopy, microscopy, and more.

Multimodal learning offers a powerful framework~\citep{li2021align, li2022blip, li2023blip}, and has been used to integrate these complementary sources of materials characterization information -- to build richer, more expressive representations of materials~\citep{ock2024unimat, mirza2025matbind, moro2025multimodal, lee2022powder, wang2024multimodal}. By jointly modeling multiple modalities, these approaches can capture correlations that are inaccessible to any single input type, leading to improved prediction accuracy, better generalization, and greater robustness to noise or missing data. 

In addition to the information inherently available in heterogeneous materials data, another major motivation for multimodal learning is the opportunity to leverage unsupervised training. While supervised models have benefited from simulated labels, generating these labels -- such as formation energy or band structure via first principles theories -- is computationally expensive and often impractical at scale. In experimental settings, labeled data is even more limited or unavailable. Multimodal frameworks enable the use of self-supervised learning objectives that exploit the natural alignment between modalities to learn meaningful representations without requiring labels. 

In this work, we present \textbf{X}RD $\boldsymbol{\times}$-\textbf{a}ttention \textbf{C}omposition \textbf{T}ransformer-\textbf{NN} (XxaCT-NN), multimodal models that do not require crystal structure information and instead operate on experimentally accessible inputs: XRD and elemental composition. We demonstrate that a cross-attention-based bimodal architecture effectively integrates these modalities to learn more expressive and transferable representations. We introduce masked XRD modeling (MXM) and explore its application to unsupervised pretraining, finding that MXM with and without contrastive alignment accelerates downstream training convergence and enhances performance. Finally, we evaluate the effect of data scale and show that multimodal models benefit proportionately from larger datasets. Together, these results demonstrate a state-of-the-art (SOTA), structure-free approach for integrating XRD and composition, offering a promising path toward foundation models grounded in experimentally available data, and potentially enabling materials discovery pipelines that bypass the labor-intensive characterization inversion step.

\section{Related work}

Multimodal learning has gained traction in materials science as a strategy for integrating diverse data sources -- such as composition, structure, and characterization -- to enhance predictive modeling. As a prerequisite, multimodality's success relies on high-quality unimodal encoder design. For the composition modality, CrabNet \citep{wang2021compositionally} and Roost \citep{goodall2020predicting} have demonstrated that self-attention architectures can effectively learn from composition alone, without relying on structural inputs. For the XRD modality, Powder XRD Pattern Is All You Need (PXRDPIAYN) \citep{lee2022powder} examines both convolutional and transformer-based architecture for understanding XRD, with the convolutional networks outperforming transformers at the data regime of the study (189,476 ICSD \citep{zagorac2019recent} entries and 139,027 Materials Project (MP) \citep{jain2013commentary} entries).

Fusion and alignment strategies have been used to improve property prediction in materials science, primarily incorporating the crystal structure modality. In COSNet \citep{wang2024multimodal} and MatFusion \citep{wan2025matfusion}, cross-attention mechanisms are used to fuse modalities without self-supervised pretraining. MatBind \citep{mirza2025matbind} and MultiMat \citep{moro2025multimodal} both take multiple materials-related modalities and use contrastive alignment among them to align the embeddings. Multimodal models have also been used to develop new maps of the materials discovery space \citep{suzuki2022selfsupervised}. While these models achieve strong results, their dependence on crystal structure limits their applicability in experimental contexts. 

Multimodality work has also reported specifically incorporating XRD and composition (without crystal structure). The XRD-composition bimodal PXRDPIAYN model ~\citep{lee2022powder} is used as a baseline in this work. Another example is UniMat~\citep{ock2024unimat}, which explores fusion and Align before Fuse (ALBEF) \citep{li2021align} on the smaller MP20 dataset (45K entries) \citep{xie2021crystal}, using contrastive alignment as the sole pretraining objective. Both works implement fusion via concatenation.

In contrast, the present work introduces a scalable cross-attention-based fusion architecture that enables dynamic, learnable interaction between composition and XRD of materials in the Alexandria dataset \citep{schmidt2023machinelearningassisted}, which is more than an order of magnitude larger than the commonly used MP dataset. 

The pretraining objectives for prior work in the materials science domain  have included contrastive alignment across different modalities (as mentioned above), or, specifically when linking to the language modality, masked language modeling (MLM) \citep{trewartha2022quantifying}. While MLM has been translated to non-language modalities for regression tasks, such as time series data \citep{dong2023simmtm}, it has not previously been adapted for materials characterization data.

\section{Method}
\label{sec:method}
\subsection{Model Architecture}
\label{sec:model_archictecture}

\begin{figure}[h!]
    \centering
    \includegraphics[width=0.5\linewidth]{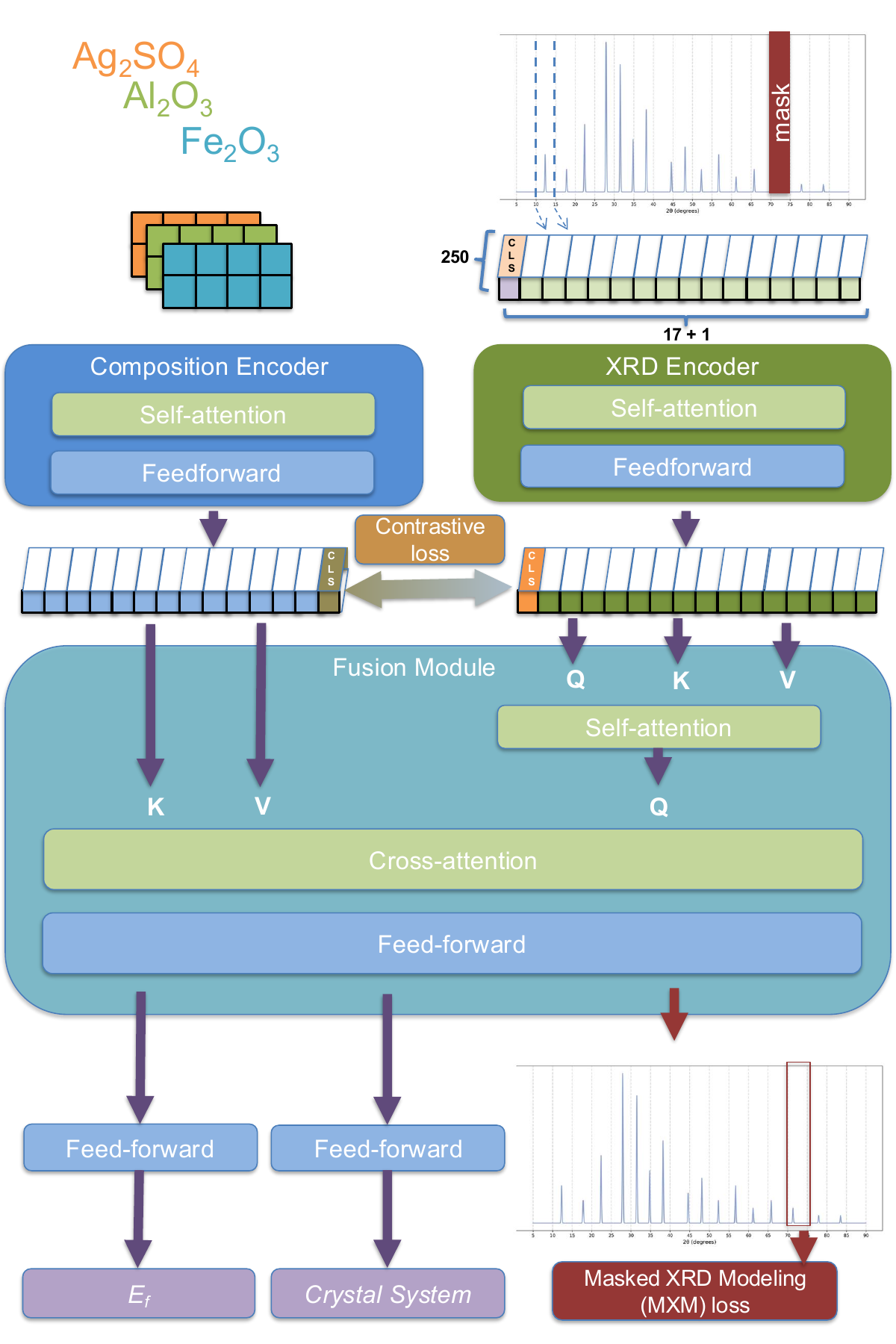} 
    \caption{Schematic illustration of \model, our proposed multimodal framework. \model consists of separate self-attention encoders for composition and XRD inputs, each producing modality-specific embeddings. 
    The embeddings are fused via a cross-attention module and jointly updated for downstream tasks, including formation energy regression and crystal system classification.
    Pretraining options include masked XRD modeling (MXM) and/or contrastive alignment.}
    \label{fig:model}
\end{figure}

The key components of our model, \textbf{X}RD $\boldsymbol{\times}$-\textbf{a}ttention \textbf{C}omposition \textbf{T}ransformer-\textbf{NN} (XxaCT-NN), illustrated in Figure \ref{fig:model}, include an encoder for the composition modality, an encoder for the XRD modality, and a multimodal fusion module.

\textbf{Composition Encoder:} We use the CrabNet architecture, trained from randomly initialized weights, to encode composition. CrabNet concatenates element identities and their stoichiometric fraction to form an element-derived matrix (EDM) representation, which is then processed by a transformer encoder to obtain the composition embeddings. The composition encoder processes a composition $C$ into a sequence of composition embeddings $\{ c_{\text{cls}}, c_1, \dots, c_N\}$. 

\textbf{XRD Encoder:} To encode the XRD modality, we use a transformer encoder that performs self-attention on the inputs. XRD patterns are represented as $4250$-dimensional vectors, which are reshaped to 17 tokens of 250 dimensions, each token being a sequence of real intensity values spanning a $2\theta$ range of $5^\circ-90^\circ$. The XRD transformer block is 3 layers deep, with an embedding dimension of 512, 4 attention heads, and 1024 dimensions for the feedforward blocks. A learnable $\texttt{[CLS]}$ token is prepended to the sequence of XRD tokens, followed by adding randomly initialized learnable positional embedding parameters. The XRD encoder processes these tokens to obtain a sequence of XRD embeddings $\{ x_{\text{cls}}, x_1, \dots, x_N\}$\footnote{Note that $N$ is overloaded to mean the sequence length of both the composition and XRD modalities although in our experiments, these are of different lengths.}. 

\textbf{Multimodal Fusion Module:} The sequence embeddings of both modalities are fed to the multimodal fusion module. We use a transformer decoder to fuse the sequence of embeddings from the composition and XRD modalities, where the composition embeddings provide key-value pairs, and the XRD embeddings provides the query for the cross-attention mechanism. This fusion module is a 12-layer deep transformer with 768 embedding dimensions, 12 attention heads, and 2048 dimensions for the feedforward blocks.

\subsection{Pretraining Objectives}
\label{sec:pretraining_obj}
Self-supervised pretraining on large-scale datasets has become a popular paradigm in representation learning, demonstrating benefits such as improved downstream performance and faster convergence~\citep{balestriero2023cookbook}. In multimodal contexts, contrastive learning (e.g., CLIP~\citep{radford2021learning}) and masked modeling (e.g., BERT~\citep{devlin2019bert}, BEiT~\citep{bao2021beit}) have emerged as particularly effective for learning aligned and transferable representations. Motivated by these advances, we explore two self-supervised objectives tailored to our bimodal architecture: contrastive learning between composition and XRD modalities, and MXM.

\textbf{Composition-XRD Contrastive Learning} Given a batch size $B$ of composition and XRD inputs, contrastive learning aims to maximize the similarity between the $B$ pairs of composition-XRD embeddings ($e_c$ and $e_x$, respectively). At the same time, the distances between the $B(B-1)$ unpaired composition-XRD embeddings are maximized. This provides for a self-supervised task to \textbf{(i)} learn a robust encoder for each modality, and \textbf{(ii)} align the composition-XRD embedding space before fusion. Following CLIP, we compute the softmax-normalized composition-to-XRD and XRD-to-composition similarity as shown in Equation~\ref{eq:similarity}.

\begin{equation}
p^{\text{c2x}}_b(C) = \frac{\exp(s(C, X_b)/\tau)}{\sum_{b=1}^B \exp(s(C, X_b)/\tau)}, \quad
p^{\text{x2c}}_b(X) = \frac{\exp(s(X, C_b)/\tau)}{\sum_{b=1}^B \exp(s(X, C_b)/\tau)} \label{eq:similarity}
\end{equation}

where $s(C, X)$ is the cosine similarity between the embeddings $\frac{c^T_\text{cls}x_\text{cls}}{||c_\text{cls}|| \cdot ||x_\text{cls}||}$ and $\tau$ is a learnable temperature parameter initialized to $0.07$. If $\mathbf{y}^\text{c2x}(C)$ and $\mathbf{y}^\text{x2c}(X)$ denote the ground truth one-hot similarity vector where matching pairs have a probability of $1$ and unmatched pairs have a probability of $0$, the symmetric contrastive loss is given by Equation~\ref{eq:contrastive_loss}.

\begin{equation}
\mathcal{L}_{\text{cont}} = \frac{1}{2} \mathbb{E}_{(C,X) \sim \mathcal{D}} \left[ \text{H}(\mathbf{y}^{\text{c2x}}(C), \mathbf{p}^{\text{c2x}}(C)) + \text{H}(\mathbf{y}^{\text{x2c}}(X), \mathbf{p}^{\text{x2c}}(X)) \right]
\label{eq:contrastive_loss}
\end{equation}

where $\mathcal{D}$ is the composition-XRD dataset, $H(\mathbf{y}, \mathbf{p})$ is the cross entropy of the distribution $\mathbf{p}$ under one-hot targets $\mathbf{y}$.

\textbf{Masked XRD Modeling (MXM):} To enhance the model's capacity to learn patterns in the XRD modality, we introduce a masked modeling objective inspired by masked language modeling (MLM). During MXM pretraining, 5\% of input XRD tokens are randomly replaced with a special $\texttt{[MASK]}$ token. The model receives the complete composition embedding and the masked XRD sequence, and is trained to reconstruct the masked region from the fused embedding. This encourages the model to learn localized patterns within each segment -- such as peak positions and shapes -- which are critical for distinguishing materials. A key difference, however, is that MXM objective is a regression task (as opposed to classification for MLM).

 Let $\{ \bar{f}_\text{cls}, \bar{f}_1, \dots, \bar{f}_N\}$ be the fused embeddings of the masked input tokens $(\bar{x}_1, \dots, \bar{x}_N)$ with each $\bar{x_i} \in \mathbb{R}^{250}$, where the original inputs $(x_1, \dots, x_N)$ are masked according to a randomly sampled binary vector $\mathbf{m} \in \{0, 1\}^N$. That is, $\bar{x}_i = \texttt{[MASK]}$ if $m_i = 1$ and $\bar{x}_i = x_i$ otherwise. Then the masked XRD modeling objective per sample is defined as the reconstruction loss over the masked regions is given by Equation~\ref{eq:mxm}, where $\{\hat{x}_1, \dots, \hat{x}_N\} = \text{ReLU}(\text{Linear}((\bar{f}_1, \dots, \bar{f}_N))$.

 \begin{equation}
     \mathcal{L}_{\text{MXM}} = \mathbb{E}_{(C, \bar{X}) \sim \mathcal{D}} \bigg[ \frac{\sum_{i=1}^{N} m_i \cdot \text{MSE}(\hat{x}_i, x_i)}{\sum_{i=1}^{N} m_i} \bigg]\label{eq:mxm}
 \end{equation}

\subsection{Data}
\label{sec:3.3}

\textbf{Preparation} The multimodal dataset used in our study is built on Alexandria 3D 2024.12.14~\citep{schmidt2023machinelearningassisted} (CC-BY 4.0 License), which includes density-functional theory (DFT) energies and crystal structures for over 5 million inorganic compounds. For each entry, the crystal structure is processed with Pymatgen (pymatgen==2024.11.13, MIT License) \citep{ong2013python}  to produce the elemental composition and simulated XRD stick patterns corresponding to Cu-K$\alpha$ radiation. With the experimental use case in mind, we perform Gaussian smearing of the XRD stick pattern ($\sigma=0.1$) to account for peak broadening, normalize intensities to $[0, 100]$, and represent XRD across a uniform $2\theta$ range of $5^\circ-90^\circ$ and spacing $0.02^\circ$ that corresponds to a $4250$-dimensional vector. In this data processing pipeline, the smearing of the stick patterns is the most expensive step. We randomly shuffle the entire dataset and split the dataset into $4,554,752$ training entries ($\approx90\%$) and $\approx491,520$ test entries ($\approx10\%$).

\textbf{Targets} The Alexandria dataset provides a variety of material properties, including formation energy and indirect band gaps. Additionally, we compute symmetry information such as crystal system and space group number using Pymatgen and include them as targets for training and evaluation.

\subsection{Experiment Details}
\label{sec:3.4}
\textbf{Single-target training} Depending on the property, each task can either be a regression task (e.g., formation energy per atom, band gap) or a classification task (e.g., crystal system, space group number). The model is trained using either mean squared error (MSE) loss for regression tasks or cross-entropy loss for classification tasks. Single-target training is only reported in Section~\ref{sec:4.1}. 

% The code for our experiments will be submitted as a zip file in the supplementary material.

\textbf{Multi-target training} In the multi-target setting, we employ a uniform task sampling strategy that selects a task per example in the batch and retrieves a corresponding training example for that task. This implicitly regularizes training by preventing the model from conditioning on task identity during encoding. While the Alexandria dataset is balanced, this strategy ensures robustness to task imbalance in broader applications. Wherever prediction of multiple targets are concerned, the same representation from the fusion module is passed through target-specific linear prediction heads. If not otherwise noted, experiments in this work are multi-target.

\label{sec:hyperparameters}
\textbf{Model sizes and hyperparameters} Our best performing model consists of a composition encoder with 9.7M parameters, XRD encoder with 6.8M parameters, and the fusion module with 94.5M parameters. In all our transformer blocks, we use a dropout of $0.1$. Our pretraining runs are for 100 epochs, and while training on target properties, we use 50 epochs. Similar to CLIP, we use a large batch size of 32,768 on 8 NVIDIA H100 (80G MEM) GPUs along with PyTorch's automatic mixed precision training~\citep{paszke2019pytorch, micikevicius2017mixed} and gradient checkpointing~\citep{chen2016training} to save memory. We use the AdamW~\citep{loshchilov2018decoupled} optimizer with $\beta_1=0.9$, $\beta_2=0.98$ and a weight decay of $0.01$. We perform a linear warmup from $10^{-6}$ to a peak learning rate of $5e^{-4}$ over $10\%$ of the training duration, followed by cosine decay to $10^{-7}$ over the remaining epochs. Since these hyperparameters were stable and did not cause gradient explosions, we do not use any gradient clipping. 

\section{Results}
\subsection{\model models outperform both unimodal and bimodal baselines}
\label{sec:4.1}
We assess predictive performance using mean absolute error (MAE) for formation energy ($E_f$) and accuracy for crystal system classification (Table~\ref{tab:alexandria-results}). For the unimodal composition baseline, we use CrabNet. For unimodal XRD, we use a transformer-based model and the CNN-based FCN model from PXRDPIAYN. Results align with materials science intuition: composition-based CrabNet yields stronger $E_f$ prediction (MAE: 131 meV/atom) but weaker symmetry classification (acc: 67.8\%), while XRD models achieve the opposite ($E_f$ MAE: 420.7, crystal system acc: 92.3\%). These results validate the quality of each unimodal model. 

For our bimodal model, we use the CrabNet architecture to embed composition, and transformer for XRD, then fuse the unimodal embeddings via a cross-attention module. The choice of transformer for XRD is based on the nominal difference between transformer vs FCN based encoder and the ease of cross-attention implementation. The resulting bimodal model achieves a substantial performance gain—$E_f$ MAE of 28.2 meV/atom and crystal system accuracy of 97.2\%, outperforming both unimodal baselines. We then freeze the encoders and the fusion module and only allow the regression layers to be updated when transferred to learn an unseen target (the band gap). The bimodal \model achieves MAEs of 0.063 eV and 0.139 eV when initialized from pretraining on $E_f$ and crystal systems, respectively. In contrast, the best unimodal result (0.191 eV) comes from the transformer-based XRD model pretrained on crystal system, $\approx$30\% worse than that of the \model counterpart. 

For comparison with prior SOTA, we reproduce the best-reported bimodal model integrating XRD and composition from PXRDPIAYN, which uses a combination of convolutional and MLP architectures. When trained on the same dataset, this model achieves a formation energy MAE of 147.6 meV/atom, over 7× higher than the performance of our proposed approach. Notably, our best-achieved $E_f$ MAE of 28.2 meV/atom approaches the SOTA performance (16 meV/atom) of structure-based GNNs trained on the same Alexandria dataset, despite using no crystal structure input.

We also explore multi-task training with both $E_f$ and crystal system as targets. While this setting slightly degrades $E_f$ performance (MAE: 40.4 meV/atom), symmetry classification remains stable, suggesting possible trade-offs in joint optimization given a fixed model size.

\begin{table}[ht]
\centering
\scriptsize
\caption{Performance on the Alexandria dataset. Fusion improves accuracy and MAE without structural input, approaching structure-based GNN performance.}
\label{tab:alexandria-results}
\begin{tabularx}{\linewidth}{c c c c c c}
\toprule
    \textbf{Type} & \textbf{Model} & \textbf{Input Modality} & \makecell{MAE $\downarrow$ \\($E_f$, meV/atom)} & \makecell{Accuracy $\uparrow$ \\(Crystal Sys., \%)} & \makecell{Transfer MAE $\downarrow$\\(Band Gap, eV)\\$E_f$ $\mid$ Crystal Sys.}\\
    \midrule
    \multirow{4}{*}{Bimodal}
    & PXRDPIAYN (FCN + MLP)  & XRD + Comp.      & 147.6   & 92.5 & $-$ \\

    & \model (single-task)   & XRD + Comp.      & \highlight{28.2}   & 96.8 & 0.063 $\mid$ 0.1387 \\
    & \model (multi-task)& XRD + Comp. & 40.4    & \highlight{97.2}  & 0.083 \\
    \midrule
    \multirow{3}{*}{Unimodal}
    & CrabNet            & Comp.      & 131   & 67.8  & 0.233 \\
    % & CrabNet            & Comp.      & 131   & 67.8  & $-$ \\
    & PXRDPIAYN (FCN) & XRD              & 420.7  & 92.3  & $-$ \\
    & XRD (Transformer)  & XRD              & 422 &  92.1  & 0.200 $\mid$ 0.191 \\
    \midrule
    Reference & ALIGNN~\citep{schmidt2024improving}        & Structure        & 16 (on different splits) & $-$   & $-$ \\
\bottomrule
\end{tabularx}
\end{table}

\subsection{Unsupervised pretraining accelerates convergence}
\label{sec:unsupervised_pretraining_result}
We investigate the impact of pretraining on training efficiency and final performance. Specifically, we evaluate three pretraining strategies for the bimodal model: (1) contrastive, (2) MXM, and (3) contrastive + MXM. These are applied in a self-supervised setting using unlabeled data prior to supervised fine-tuning on $E_f$ and crystal system prediction.

\begin{table}
\centering
\small
\caption{Impact of pretraining on \model model performance and convergence progress. Lower MAE and higher accuracy indicate better performance.}
\label{tab:pretrain-results}
\begin{tabularx}{\textwidth}{lcc}
\toprule
\textbf{Pretraining Strategy} & \makecell{\textbf{MAE} $\downarrow$ \\ ($E_f$, meV) \\ (5\%, 15\%, 25\% training, Best)} & \makecell{\textbf{Accuracy} (\%) $\uparrow$ \\ (Crystal System) \\ (5\%, 15\%, 25\% training, Best)} \\
\midrule
Without pre-training (baseline) & 167, 97, 81, 45.7     & 83.4, 93.6, 95.8, 97.19 \\
Contrastive                     & 105, 87, 76, 44.49    & 93.9, 95.9, 96.4, 97.21 \\
Masked XRD Modeling (MXM)      & 101, \textbf{79}, 72, \textbf{43.48}    & 94.7, \textbf{96.3}, 96.5, 97.21 \\
\textbf{Contrastive + MXM}     & \textbf{91}, 80, \textbf{68}, 43.82    & \textbf{95.2}, 96.2, \textbf{96.7}, \textbf{97.32} \\
\bottomrule
\end{tabularx}
\end{table}

 As shown in Table \ref{tab:pretrain-results}, all three pretraining approaches—contrastive, MXM, and their combination—consistently improve prediction accuracy and reduce formation energy MAE compared to training from scratch. Without pretraining, the model achieves a best MAE of 45.7 meV and 97.19\% accuracy. With contrastive or MXM pretraining, MAE improves to 44.49 meV or 43.48 meV, respectively, with corresponding accuracy gains. By combining both strategies, the model achieves a MAE of 43.82 meV and highest accuracy of 97.32\%.

To assess training efficiency, we compare all pretraining strategies using checkpointed models at 5\%, 15\%, and 25\% of training. Across all levels, pretraining consistently improves both tasks over the baseline. At 5\% training, the combined contrastive + MXM model achieves the largest gains—reducing MAE by 76 meV and improving accuracy by +11.8\%. MXM alone yields similar improvements (-66 meV, +11.3\%), while contrastive pretraining results in -62\,meV MAE and +10.5\% accuracy. These trends persist at larger data fractions: at 15\% and 25\% training. If we use the baseline accuracy at 25\% training as threshold MAE $\leq$ 0.081 and accuracy $\geq$ 95.8\%, the contrastive + MXM pretrained model reaches this level within 3,000 iterations, while the baseline requires over 12,000 iterations, yielding a 4.2$\times$ speed-up. MXM-only and contrastive-only models also reach the threshold faster than the baseline, at approximately 2,300 and 3,500 iterations, corresponding to 1.8$\times$ and 1.2$\times$ speed-ups, respectively. 

Between the individual strategies, MXM pretraining provides a larger improvement than contrastive loss, particularly in reducing the MAE of $E_f$ (43.48 vs 44.49 meV/atom). While both approaches yield similar accuracy on crystal system classification, the MXM-pretrained model exhibits more distinct symmetry-aligned clusters in the latent space according to Figure~\ref{fig:latent-space} (Silhouette score 0.45 vs. 0.50). Additionally, the model pretrained with MXM presents a larger speed-up (1.8$\times$ vs 1.2$\times$) in regression, with noticeably smoother and more stable convergence curves (Figure~\ref{fig:convergence}). This can be attributed to the fact that the MXM objective updates both the fusion module and the encoders, allowing the model to jointly refine intra- and inter-modality representations. In contrast, the contrastive loss primarily regularizes the encoders and does not directly influence the fusion block.

\subsection{\model models learn more physically meaningful representations}
\label{sec:4.3}

\begin{figure}[ht]
    \centering
    \includegraphics[width=\linewidth]{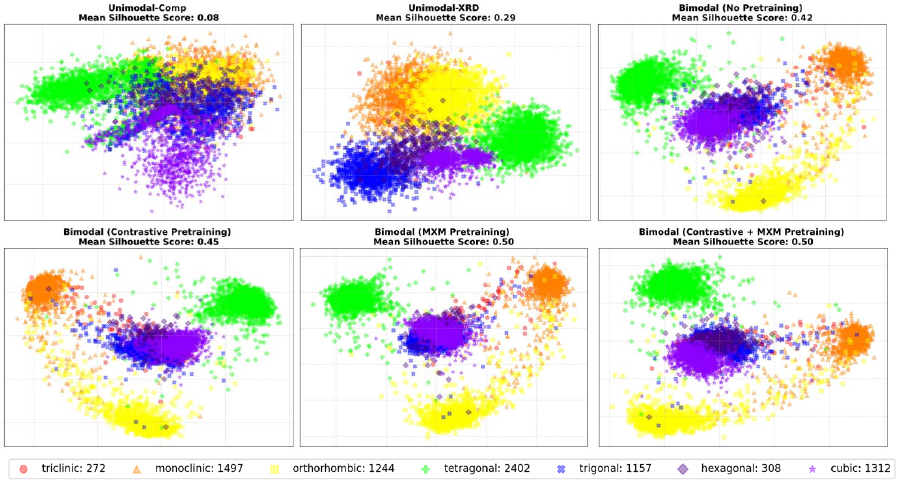}  
    \caption{PCA visualizations of learned latent embeddings colored by crystal system. \textbf{Top row}: unimodal models using composition-only (left, silhouette: 0.08), XRD-only (center, 0.29), and bimodal (right, no pretraining, 0.42). \textbf{Bottom row}: bimodal models with contrastive pretraining (0.45), MXM pretraining (0.50), and contrastive + MXM pretraining (0.50). Multimodal fusion significantly improves cluster separation, and pretraining further enhances the structure of the latent space.}
    \label{fig:latent-space}
\end{figure}
To better understand the impact of multimodal fusion on representation learning, we analyze the latent space clusters produced by unimodal and bimodal models. Figure~\ref{fig:latent-space} visualizes 2D PCA projections of embeddings learned from composition-only, XRD-only, and various bimodal configurations. Each point is colored by its ground-truth crystal system.

Qualitatively, the unimodal composition model produces poorly clustered embeddings (mean silhouette score: 0.08), while the XRD model exhibits slightly better separation (0.29). In contrast, bimodal models show substantially improved clustering aligned with crystal system labels. Without pretraining, the fused model already achieves a silhouette score of 0.42. With either contrastive pretraining or MXM pretraining, the score improves further to 0.45 and 0.50, respectively. When pretrained with both losses, the mean silhouette score maintained at 0.50. We also performed more thorough examination of the accuracy for individual crystal systems (Table \ref{tab:cs_silh_score}), through which we find the trend holds as the accuracies for majority of the systems present the same order as the mean scores with the exception of cubic. The Transformer-XRD model presents a significant bias towards more symmetric systems, especially cubic, whereas the bimodal models distinguish all classes more evenly, which is a sign of a more comprehensive, and physically meaningful representation. 

% To better understand how each modality contributes to prediction, we analyzed the learned attention maps from the cross-attention fusion block (see Fig. X, \jith{get the cross attention matrix}). We observe that crystal system classification predominantly attends to XRD features, while formation energy prediction relies more on composition. This aligns well with domain knowledge and highlights the model’s ability to adaptively leverage modality-specific information.

\subsection{Larger datasets favor multimodal models}

We examine how dataset scale influences predictive performance in multimodal materials models. First, we evaluate the impact of training set size on a fixed architecture. We compare a unimodal model trained on composition alone with a bimodal model combining composition and XRD, across dataset sizes ranging from 1M to 4.5M examples.

\label{sec:4.4}
\begin{wrapfigure}{r}{0.48\textwidth}
    \vspace{-2.7cm}
    \centering
    \includegraphics[width=\linewidth]{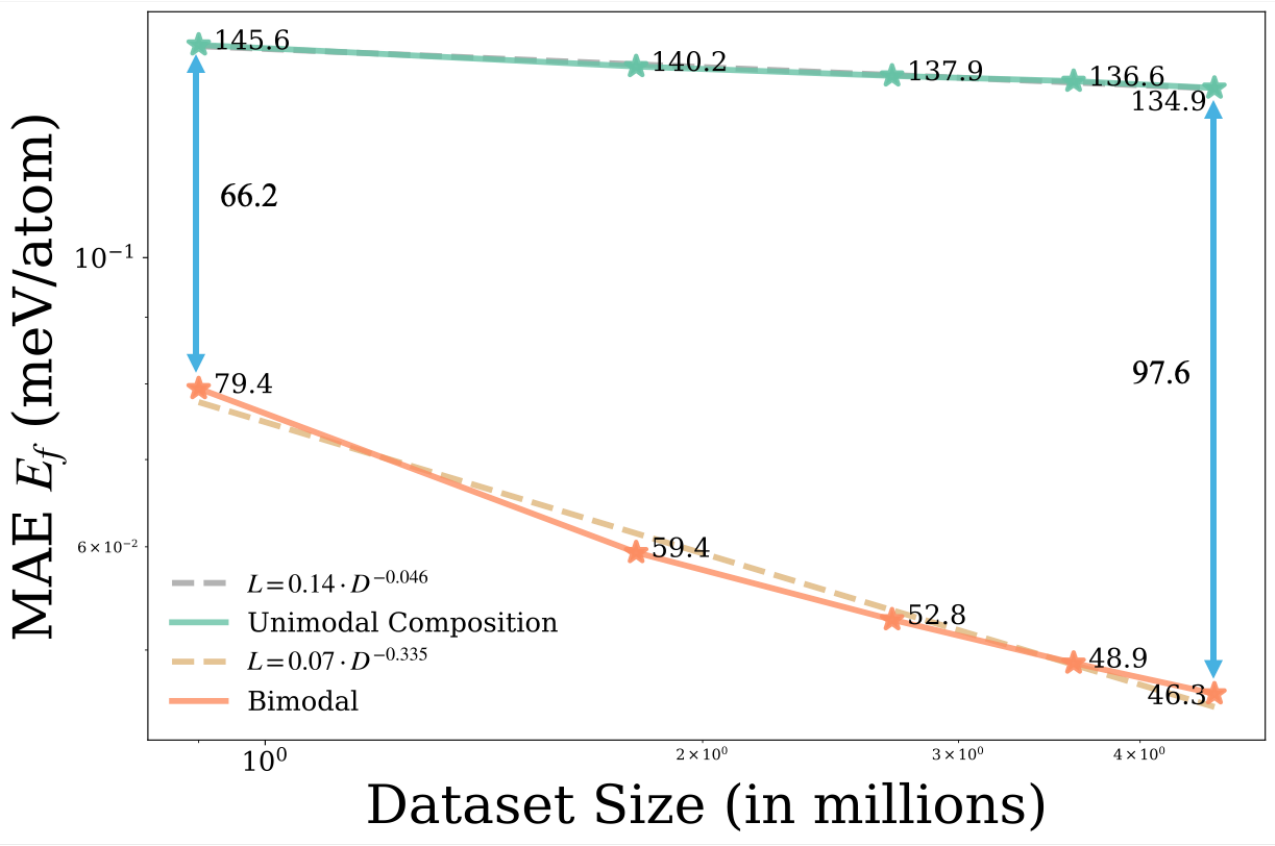}
    \caption{ Test set performance as a function of dataset size. \model (orange) has a more negative scaling exponent compared to the unimodal model (green). The performance gap widens from 66.2 meV to 97.6 meV as data increases from 1M to 4.5M, demonstrating that multimodal models benefit more from scaling.}

    % \caption{\small \textbf{Test set performance vs. dataset size:} \model (orange) scales better than the unimodal model (green), with the performance gap growing from 66.2 to 97.6 meV as data increases from 1M to 4.5M.}
    \label{fig:scaling}
\end{wrapfigure}

As shown in Figure~\ref{fig:scaling}, the unimodal model exhibits limited performance improvement with scale, following a weak power-law fit: \( L = 0.14 \cdot D^{-0.046} \), where \( L \) is MAE and \( D \) is dataset size. In contrast, the bimodal model follows a significantly stronger scaling trend: \( L = 0.07 \cdot D^{-0.335} \). At 1 million examples, the bimodal model already outperforms the unimodal baseline by 66.2 meV. This gap increases to ~97.6 meV at 4.5 million samples, indicating that multimodal models not only yield better baseline accuracy, but also exhibit amplified returns with more data. Note that for all experiments in Figure~\ref{fig:scaling}, the models is evaluated on the same test set of 491,520 entries.

% Next, we assess how model size interacts with pretraining. In Figure~\ref{fig:scaling}, we plot MAE versus parameter count for various architectures, comparing models trained from scratch to those initialized with contrastive pretraining. Across all model scales—ranging from ~40M to 125M parameters—contrastive pretraining consistently improves performance. Larger models especially benefit, with pretraining providing up to 15 meV reduction in MAE.

\section{Discussion}
\label{discussion}

Our focus on structure-agnostic machine learning approaches is motivated by situations where explicit crystal structures are unknown or too complex to be solvable in practice. This assumption is not to say that crystal structure is not useful in experiment-forward approaches. Rather, we believe that structure-agnostic experiments of machine learning may teach us when and how to use crystal structure as an inductive bias. 

This work uses the composition and XRD modalities, which have been important in the foundations of materials science. Early practitioners knew only of the composition of the materials investigated and the interaction of their materials with X-rays. By recognizing that (a) the materials were diffracting the X-rays, (b) that those diffraction patterns corresponded to planes in the crystal structure, and (c) that the observed evidence of simultaneous crystal planes evidenced the three-dimensional, periodic atomic structure, these practitioners labeled their observations with crystal structures.

Today, of course, we have the luxury of referring to large databases of crystal structures labeled using experimental data (including XRD and other techniques) and those from simulations using theories derived from quantum mechanics. However, given that crystal-structure-containing datasets are expensive to assemble/generate, and those that already exist receive a large fraction of the attention of the materials research community, we believe structure-agnostic experiments present new opportunities.

With the expectation that unlabeled but paired data may be used to make up larger materials datasets, we present MXM loss, and benchmark models with MXM and contrastive pretraining. In short, we find that pretraining objectives from the vision-language domain can be adapted to materials science domain models. Pretraining was found to accelerate training and improve the structure of the embedding space, with the biggest gains for MXM and MXM+contrastive pretraining. MXM may be further developed to reflect real-world use cases like low resolution and/or noisy XRD data.

Most prior work in multimodal materials learning has been limited to datasets containing up to hundreds of thousands of entries (\cite{jain2013commentary, zagorac2019recent}. Our results on the 5-million entry Alexandria dataset provide two key insights into the role of data scale. First, at the level of individual encoder development, we observe that performance bottlenecks are not solely due to data limitation.  According to Table \ref{tab:pretrain_comparison} and Figure \ref{fig:crystal_system_distribution}, hexagonal and triclinic are the least frequent classes, yet the accuracies achieved by all models on high symmetry hexagonal class are consistently better than that of not only triclinic, but some of more frequent classes such as orthorhombic and monoclinic. In the decreasing order of data abundance but increasing order of crystal symmetry, the accuracies increase as we go from monoclinic, to orthorhombic, to trigonal. These observations suggest that the difficulty lies more in distinguishing low-symmetry classes than in the scarcity of training data. Hence representation design with better inductive biases remain crucial, particularly for modalities like XRD. 

Secondly, when scaling both data and model capacity, we find that performance continues to improve steadily, with no indication of saturation. Our findings clearly show that the field of materials informatics has not yet reached a performance plateau. Importantly, multimodal frameworks appear especially well-suited for training at scale, as they can flexibly integrate complementary information across modalities to better exploit large datasets.

% To address the concern that the performance gains of the bimodal model may simply result from increased parameter count, we conducted controlled experiments in which all unimodal and bimodal models were constrained to have approximately the same number of parameters. As shown in Figure Sx, the best-performing bimodal configurations consistently outperformed their unimodal counterparts at similar model sizes. This indicates that the observed improvements stem primarily from the additional information provided by multiple modalities, rather than from differences in model capacity.

\textbf{Limitations:} \label{sec:limitations} As noted earlier, the modalities used in this work—composition and XRD—are more experimentally accessible than atomic crystal structures. However, the Alexandria dataset is composed of simulated materials, and we have not yet addressed the adaptation challenge of transferring to real experimental inputs. The specific challenges are: 
(a) The current model does not explicitly account for the domain shift between simulated and experimental data. For example, on the XRD side, a more comprehensive treatment of experimental noise sources—including background signals, instrument artifacts, and peak shift—is needed.  
(b) The current input forms of the data is still analytically inferred from raw experimental measurements. For example, for composition, constituent elements and stoichiometry are analytically inferred from characterization methods such as energy-dispersive X-ray spectroscopy (EDS) or X-ray fluorescence (XRF). 

Apart from the challenge of experimental adaptation, a separate limitation is the restricted set of input modalities (XRD, composition) and tasks (crystal system, formation energy, and band gap prediction). The tasks in this work have been well-studied, with prior models already demonstrating good performance from the domain perspective and making significant gains difficult to obtain. Nevertheless, \model models demonstrate quantitative improvements, and similar multimodal models potentially can produce more significant improvements as more challenging tasks and novel modalities are incorporated.

\section{Conclusion}
\label{conclusion}
We present \model , a structure-free multimodal framework that integrates more experimentally accessible modalities, XRD and elemental composition, using a cross-attention architecture. This model achieves strong performance on formation energy and crystal system prediction without relying on crystal structural input while achieving performance on par with a SOTA structure-based model. We further show that self-supervised pretraining -- via contrastive alignment and masked XRD modeling -- improves convergence, representation quality, and scalability. Our results highlight the potential of multimodal learning to build scalable, experimentally grounded foundation models for materials science.

\bibliographystyle{unsrtnat}
\bibliography{neurips_2025}

% References follow the acknowledgments in the camera-ready paper. Use unnumbered first-level heading for
% the references. Any choice of citation style is acceptable as long as you are
% consistent. It is permissible to reduce the font size to \verb+small+ (9 point)
% when listing the references.
% Note that the Reference section does not count towards the page limit.
% \medskip

% {
% \small

% [1] Alexander, J.A.\ \& Mozer, M.C.\ (1995) Template-based algorithms for
% connectionist rule extraction. In G.\ Tesauro, D.S.\ Touretzky and T.K.\ Leen
% (eds.), {\it Advances in Neural Information Processing Systems 7},
% pp.\ 609--616. Cambridge, MA: MIT Press.

% [2] Bower, J.M.\ \& Beeman, D.\ (1995) {\it The Book of GENESIS: Exploring
%   Realistic Neural Models with the GEneral NEural SImulation System.}  New York:
% TELOS/Springer--Verlag.

% [3] Hasselmo, M.E., Schnell, E.\ \& Barkai, E.\ (1995) Dynamics of learning and
% recall at excitatory recurrent synapses and cholinergic modulation in rat
% hippocampal region CA3. {\it Journal of Neuroscience} {\bf 15}(7):5249-5262.
% }

% %%%%%%%%%%%%%%%%%%%%%%%%%%%%%%%%%%%%%%%%%%%%%%%%%%%%%%%%%%%%

\appendix

\section{Appendix}
\label{sec:appendix}

\subsection{Training and Validation Data Distribution}

\begin{figure}[htbp]
  \centering
  \subfigure[Train distribution]{
    \includegraphics[width=0.48\textwidth]{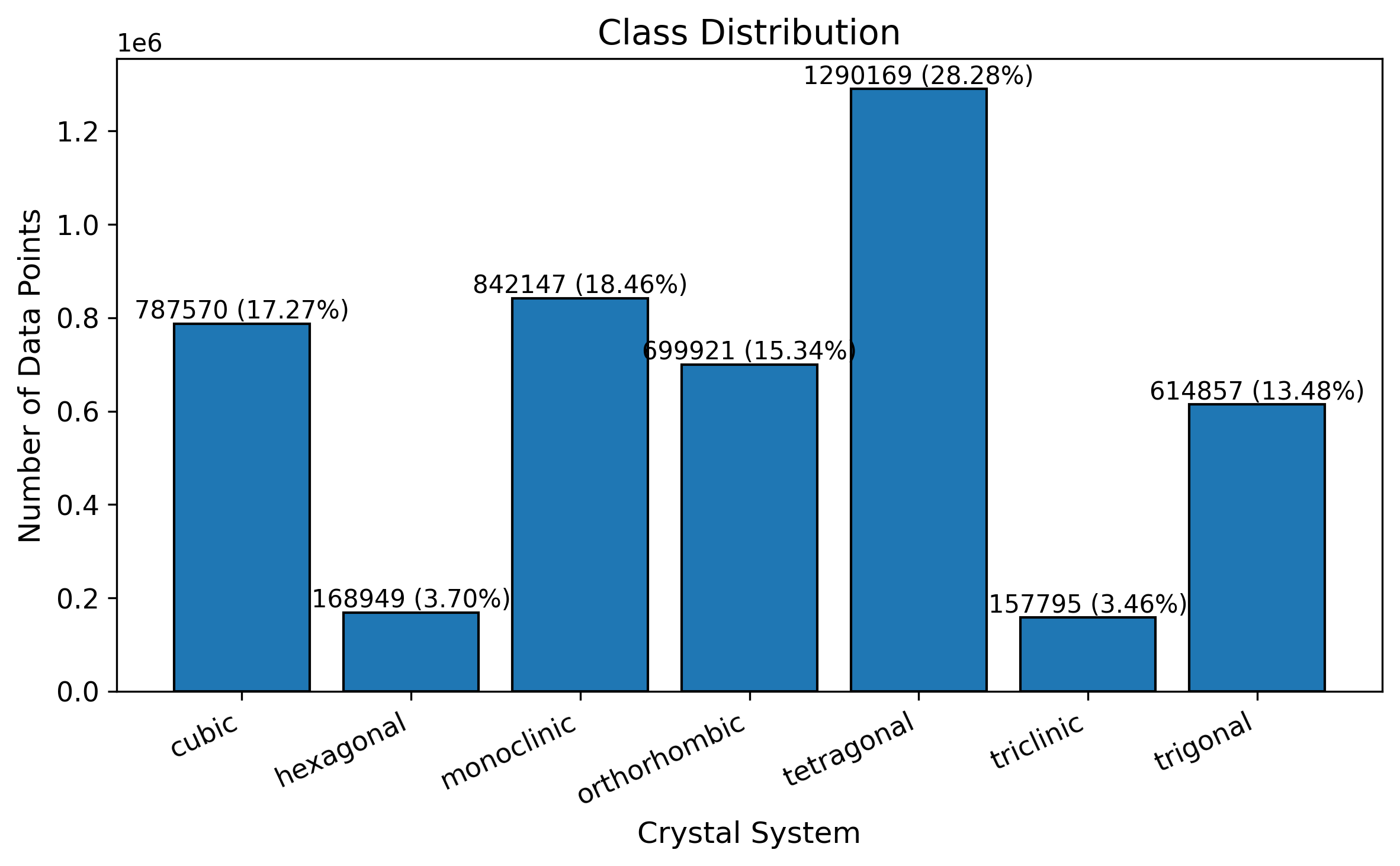}
  }
  \hfill
  \subfigure[Test distribution]{
    \includegraphics[width=0.48\textwidth]{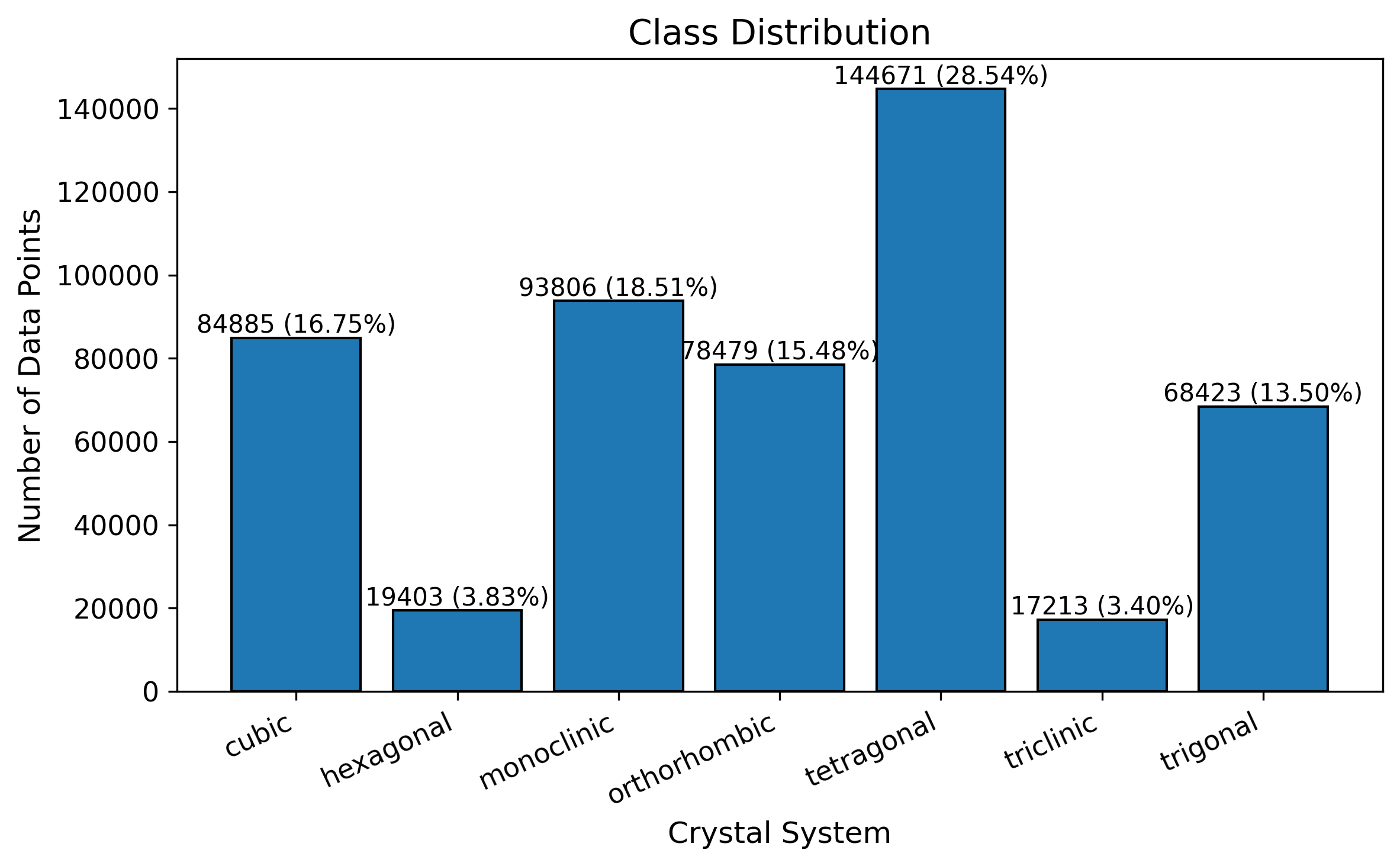}
  }
  \caption{Crystal system distribution across train and test sets.}
  \label{fig:crystal_system_distribution}
\end{figure}

Figure \ref{fig:crystal_system_distribution} shows the distribution of crystal systems in the training and validation datasets.

\begin{figure}[htbp]
  \centering
  \subfigure[Train distribution]{
    \includegraphics[width=6.5cm]{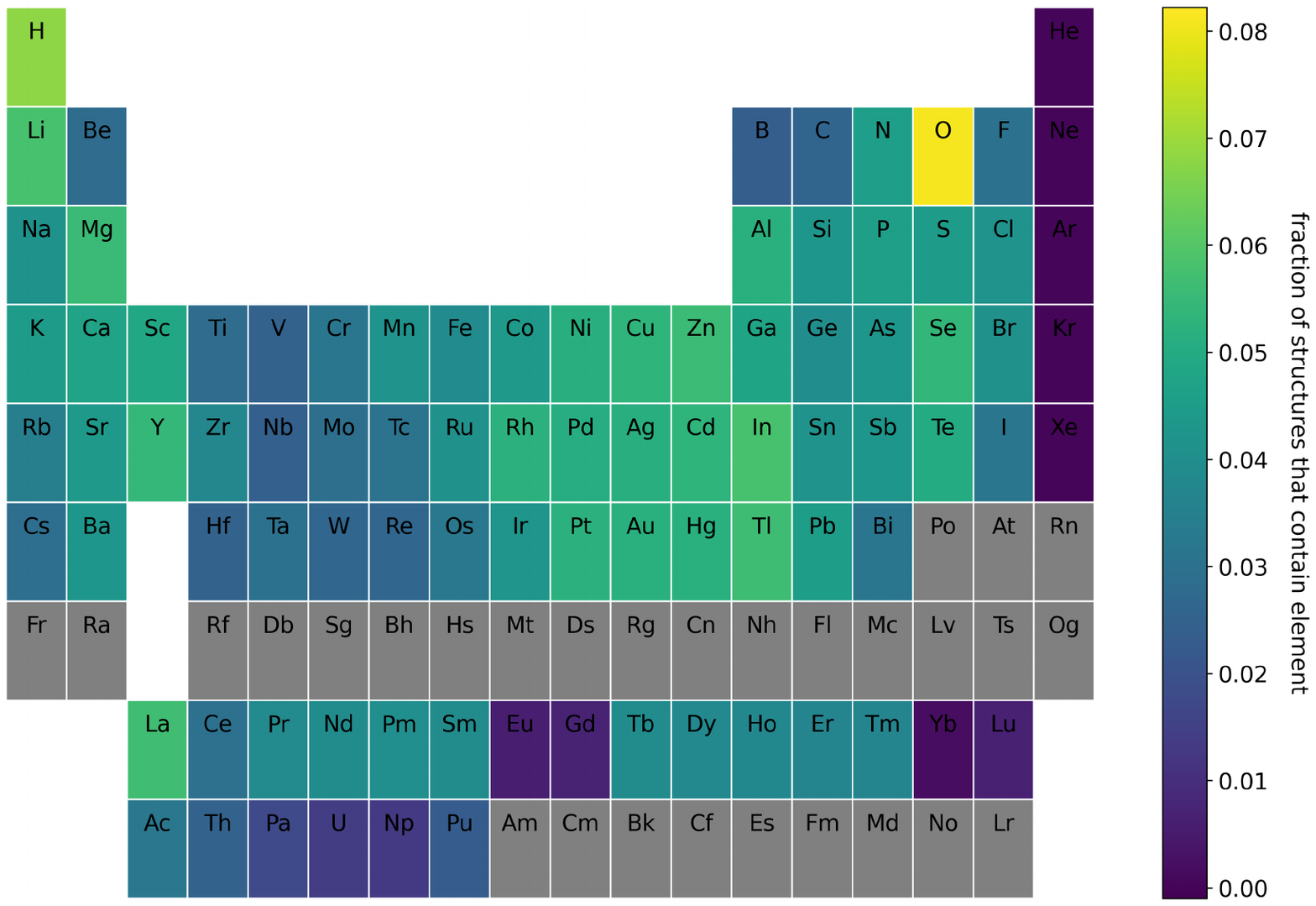}
  }
  \hfill
  \subfigure[Test distribution]{
    \includegraphics[width=6.5cm]{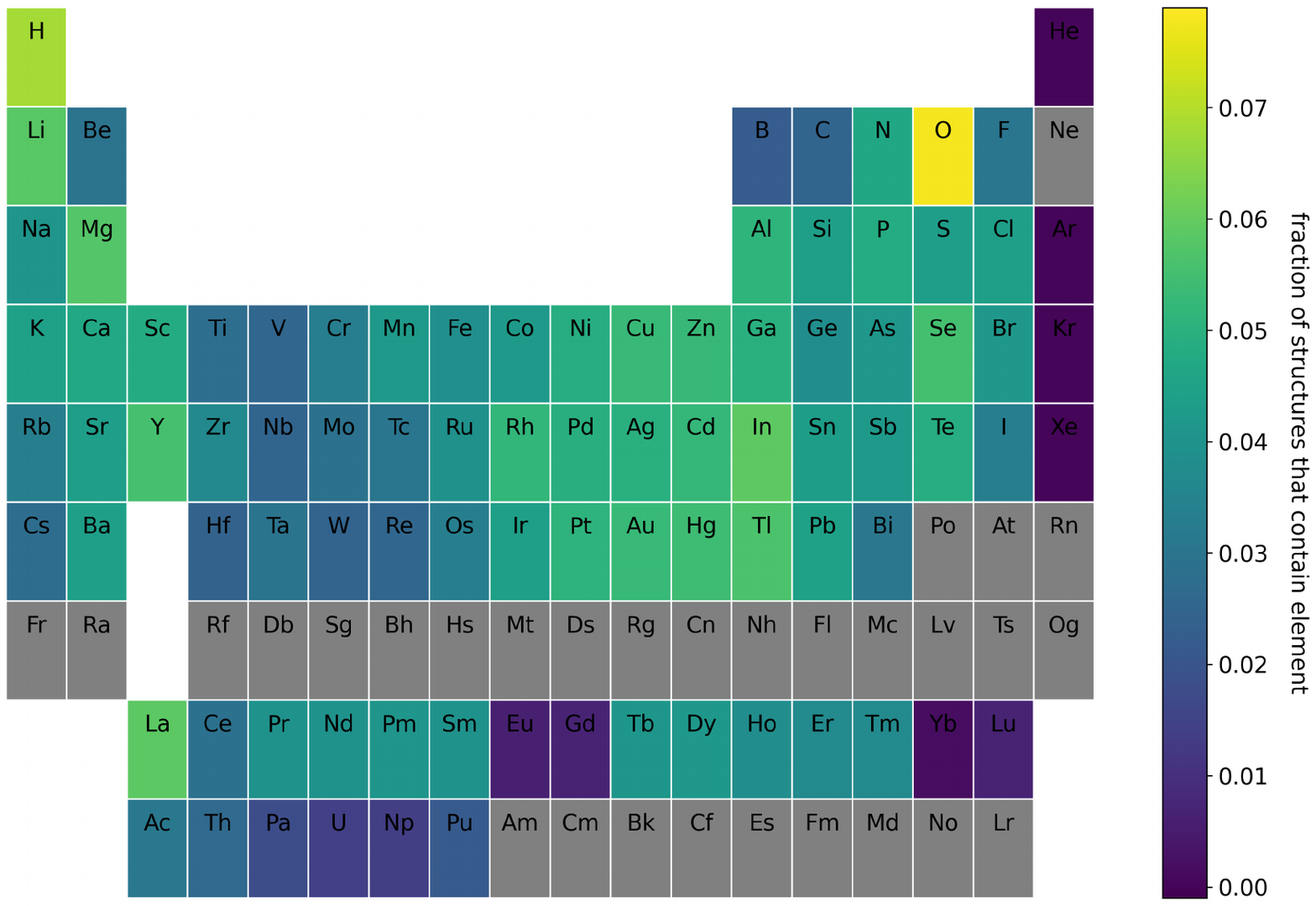}
  }
  \caption{Visualizing element distribution in Alexandria}
  \label{fig:periodic_table}
\end{figure}

Figure \ref{fig:periodic_table} shows the distribution of containing elements in the training and validation datasets.

\subsection{Benchmark training}
In order to compare our findings to previous work, we chose to retrain unimodal (FCN) and bimodal (FCN+MLP) architectures presented in the PXRDPIAYN publication and code on the Alexandria dataset. We have translated the code from Tensorflow into Pytorch and integrated it into our training flow.
In order to adapt the architecture to our x-ray diffraction spectra dimension of 4250 spectral values for a given structure, vs. an original of 8192, we make small changes to individual layer dimensions. These changes result in a small reduction of learnable parameters in the FCN + MLP model (1,814,401 in our adaptation vs 1,879,937 in the original).
We trained on a batch size of 32,768 using the AdamW optimizer. Exploring training hyperparameters, we went through a number of iterations to arrive at the results in table \ref{tab:pxrdpiayn-benchmark} for given pairs of architecture (FCN, FCN+MLP) and targets (Ef, crystal system)

\begin{table}
    \centering
    \caption {Performance of PXRDPIAYN reference models trained on single targets on Alexandria dataset}
    \label{tab:pxrdpiayn-benchmark}
    \begin{tabularx}{\linewidth}{c c c c c c}
    \toprule
    \textbf{Model} & \textbf{Input Modality} & \textbf{Target} & \textbf{Epochs} & \makecell{\textbf{MAE} $\downarrow$ \\($E_f$, meV)} & \makecell{\textbf{Accuracy} $\uparrow$ \\(Crystal System, \%)} \\
    \midrule
    FCN & XRD & $E_f$ & 50 & 422.8 & - \\
    FCN & XRD & $E_f$ & 100 & 420.7 & - \\
    FCN & XRD & Crystal System & 100 & - & 90.78 \\
    FCN & XRD & Crystal System & 200 & - & 92.29 \\
    FCN + MLP & XRD + Comp. & $E_f$ & 50 & 147.1 & - \\
    FCN + MLP & XRD + Comp. & $E_f$ & 100 & 147.6 & - \\
    FCN + MLP & XRD + Comp. & Crystal System & 50 & - & 83.2 \\
    FCN + MLP & XRD + Comp. & Crystal System & 100 & - & 89.93 \\
    FCN + MLP & XRD + Comp. & Crystal System & 200 & - & 92.51 \\
    \bottomrule
    \end{tabularx}
\end{table}

% XRD + Comp. 

\subsection{Impact of pretraining on model performance and convergence}

\begin{figure}[ht]
    \centering
    \includegraphics[width=\linewidth]{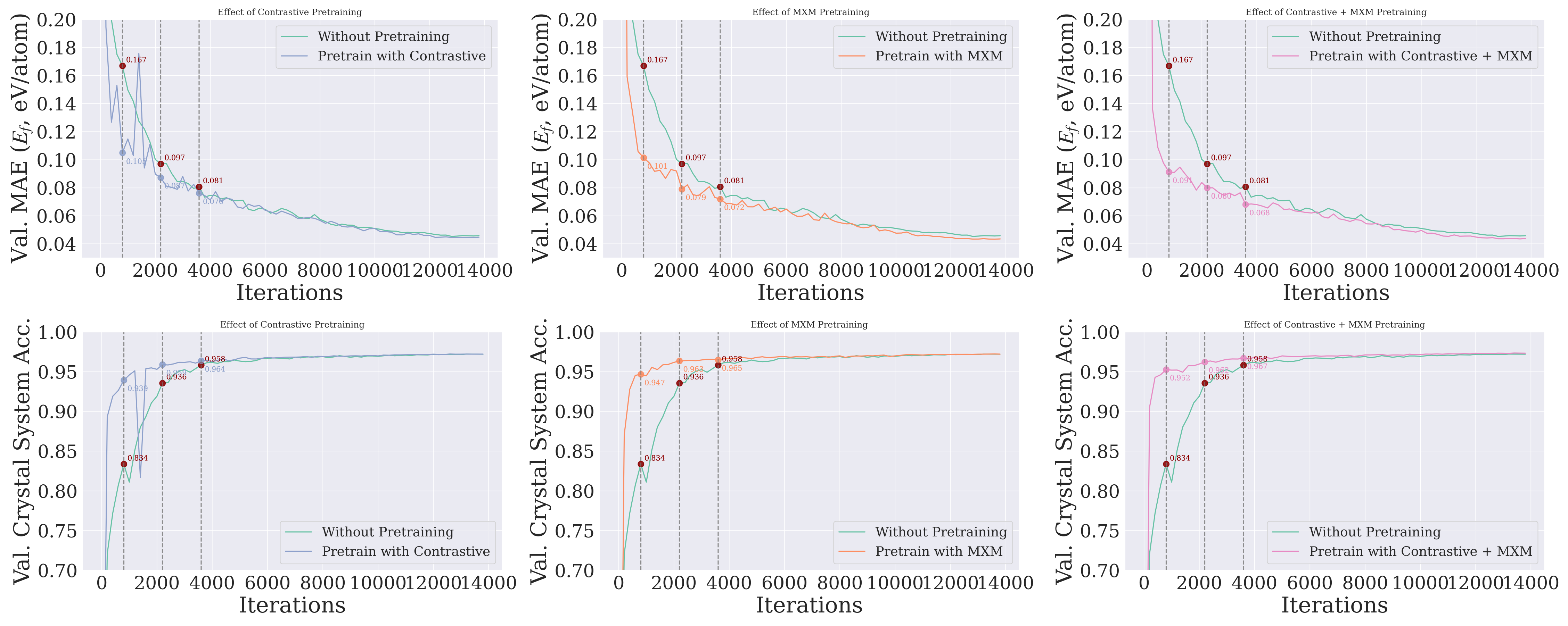}
    \caption {Effect of pretraining on downstream performance and convergence speed. Top row: validation MAE for $E_f$; bottom row: validation accuracy for crystal system classification. Each column corresponds to a different pretraining strategy—contrastive + MXM (left), contrastive only (middle), and MXM only (right) -- compared to models trained without pretraining. Pretrained models consistently achieve lower error and higher accuracy with significantly faster convergence, demonstrating the effectiveness of self-supervised objectives.}
    \label{fig:convergence}
\end{figure}

Figure~\ref{fig:convergence} illustrates compares the effect pretraining from contrastive, MXM, and contrastive + MXM objectives with respect to a bimodal model without pretraining. 

Table~\ref{tab:pretrain_comparison} lists the accuracies per crystal system on the XRD transformer model and \model, with an ablation on the pretraining objectives. In Table~\ref{tab:cs_silh_score}, we also provide the silhouette scores per crystal system for all 6 models corresponding to the 6 PCA plots shown in Figure~\ref{fig:latent-space} of the main text.

\begin{table}  
    \small
    \centering
    \caption{Accuracy (\%) across crystal systems under different model configurations. \highlight{Blue} highlights show the best performing model per crystal system. \textbf{Bolded} accuracy refers to the  model that performs best across all crystal systems.}
    \begin{tabular}{lccccc}
  
        \toprule
        \makecell{ Crystal System } & \makecell{ XRD \\ Transformer }  & \makecell{ \model \\ No pretraining }  & \makecell{ \model \\ w/o MXM }  & \makecell{ \model \\ w/o Contrastive } & \makecell{ \model } \\
        \midrule
        Triclinic     &74.31 &\highlight{88.26} & 87.51 & 87.28 & 87.87 \\
        Monoclinic    &87.15 & 94.36 & 94.15 & 94.10 & \highlight{94.48} \\
        Orthorhombic  &83.24 & 94.81 & 95.20 & 95.19 & \highlight{95.30} \\
        Tetragonal    &97.77 & 99.32 & 99.34 & 99.34 & \highlight{99.37} \\
        Trigonal      &93.95 & 98.55 & \highlight{98.63} & 98.61 & 98.62 \\
        Hexagonal     &87.32 & 96.37 & 96.33 & \highlight{96.73} & \highlight{96.73} \\
        Cubic         &99.60 & 99.81 & 99.81 & \highlight{99.84} & 99.83 \\
        \midrule
        Mean          &92.14 & 97.20 & 97.21 & 97.21 & \textbf{97.32} \\
        \bottomrule
    \end{tabular}
    \label{tab:pretrain_comparison}
\end{table}

\begin{table}[ht]
    \small
    \centering
    \caption{Silhouette scores per crystal system for visualizations in Figure~\ref{fig:latent-space}. Generally, going from no pretraining $\rightarrow$ w/o MXM $\rightarrow$ w/o contrastive $\rightarrow$ \model, we notice the silhouette scores per class consistenly increase for all crystal systems. In general, all variants of \model also exhibit better clustering as per the silhouette score, when compared to unimodal (CrabNet, XRD Transformer) variants, with the exception of the cubic class.}
    \begin{tabular}{lcccccc}
        \toprule
        \makecell{ Crystal System }
        & \makecell{ CrabNet } 
        & \makecell{ XRD \\ Transformer } 
        & \makecell{ \model \\ No pretraining } 
        & \makecell{ \model \\ w/o MXM } 
        & \makecell{ \model \\ w/o Contrastive } 
        & \makecell{ \model } \\
        \midrule
        Triclinic     & 0.07 & 0.19 & 0.33 & 0.36 & 0.37 & 0.37 \\
        Monoclinic    & -0.03 & 0.05 & 0.42 & 0.45 & 0.49 & 0.50 \\
        Orthorhombic  & 0.06 & 0.21 & 0.44 & 0.48 & 0.52 & 0.51 \\
        Tetragonal    & 0.14 & 0.38 & 0.41 & 0.43 & 0.50 & 0.49 \\
        Trigonal      & 0.06 & 0.24 & 0.45 & 0.47 & 0.55 & 0.53 \\
        Hexagonal     & 0.13 & 0.37 & 0.51 & 0.53 & 0.59 & 0.58 \\
        Cubic         & 0.14 & 0.52 & 0.39 & 0.41 & 0.47 & 0.46 \\
        \midrule
        Mean          & 0.08 & 0.29 & 0.42 & 0.45 & 0.50 & 0.50 \\
        \bottomrule
    \end{tabular}
    \label{tab:cs_silh_score}
\end{table}

% (val) Run d4yf5snw: 0.50

% (val) Run 2zg6o6fj: 0.50

% triclinic: 0.37
% trigonal: 0.53

\subsection{Execution time}

For experiments with largest model (112M parameters) on the Alexandria data set, running 50 epochs takes about 4 hours for single target runs, and 8 hours for multi-target runs. For any pretraining run with the largest model, the runtime is upto 6 hours.

\end{document}